# A Deep Reinforcement Learning-Based Charging Scheduling Approach with Augmented Lagrangian for Electric Vehicle

Guibin. Chen, *Student Member, IEEE,* Xiaoying. Shi, *Student Member, IEEE*

*Abstract*—This paper addresses the problem of optimizing charging/discharging schedules of electric vehicles (EVs) when participate in demand response (DR). As there exist uncertainties in EVs' remaining energy, arrival and departure time, and future electricity prices, it is quite difficult to make charging decisions to minimize charging cost while guarantee that the EV's battery state-of-charge (SOC) is within certain range. To handle with this dilemma, this paper formulates the EV charging scheduling problem as a constrained Markov decision process (CMDP). By synergistically combining the augmented Lagrangian method and soft actor critic algorithm, a novel safe off-policy reinforcement learning (RL) approach is proposed in this paper to solve the CMDP. The actor network is updated in a policy gradient manner with the Lagrangian value function. A double-critics network is adopted to synchronously estimate the action-value function to avoid overestimation bias. The proposed algorithm does not require strong convexity guarantee of examined problems and is sample efficient. Comprehensive numerical experiments with real-world electricity price demonstrate that our proposed algorithm can achieve high solution optimality and constraints compliance.

## I. INTRODUCTION

WITH the growing concerns on environmental issues, electric vehicles (EVs) have been considered as a promising solution to reduce carbon emission and popularly adopted worldwide in recent years [1-3]. A report by International Energy Agency (IEA) reveals that the global number of EVs reached about 11.2 million in 2020 and this number will further increase to 125 million by 2030 [4]. However, the intermittent charging demands of EVs introduce challenges to the operation of power system such as elevated load peaks, voltage deviation and increased power loss [5]. Designing the mechanism that alleviates such stress is imperative. Recently, the demand response (DR) for the EV is proposed, which aims to shape its daily demand profile and therefore flatten the total load [6-8]. When participate in DR, EVs' charging schedules can be optimized in response to the time-varying price and therefore cut down their electricity bills [9].

Nevertheless, designing real-time charging/discharging schedules for an EV is quite challenging due to the uncertainties in EV user's behavior, including the charging demand, arrival/departure time and future electricity price. To satisfy EV users' comfort, a straightforward way is to get EVs fully charged as soon as possible once they arrive, but this is low-efficient and may cause overloads on distribution network during the period of EVs' large-scale integration. The ideal scenario is to take advantage of the time-varying price and schedule a proper quantity for EVs to make charging/discharging decision, thus achieve users' comfort and cost minimization simultaneously. However, due to the absence of future information, it is difficult to optimize such multiobjective tradeoff problem.

With the capability of tackling sequential decision-making problems under uncertain environment and learning from feedback to optimize certain performance, deep reinforcement learning (DRL) has gained strong interest recently. Many researchers have adopted DRL to crack individual EV charging control problem and obtained excellent performance. In [10], the deep Q-networks learning (DQN) algorithm is adopted to solve the individual EV charging scheduling problem, which introduces the long short-term memory (LSTM) neural network to extract features from price signal. However, the DQN can only be applied in finite discrete action space. To address the continuous charging rates, [11] and [12] introduce the deep deterministic policy gradient (DDPG) to replace the DQN. Since DDPG relies heavily on manually adding noise on output action and is sensitive to hyperparameter settings, a maximum entropy DRL algorithm, the soft actor critic (SAC), is adopted by [12] to train independent learner for learning optimal charging strategy for each EV in a multi-agent framework. [14] establishes a mathematical model to quantitatively describe the EV users' anxiety on battery state-of-charge (SOC), then uses the SAC to make real-time charging decisions. In [15], the historical data on EVs dynamics is used to develop statistical pattern, which provides a general environmental setting for the DRL agent. The proximal policy optimization (PPO) is adopted to address continuous state and action, while the reward function is explicitly revised to fulfill EVs' charging demand. [16] develops an online smart charging algorithm based on asynchronous actor critic method to schedule the EV charging against uncertainties in EV charging behaviors. To minimize

G. Chen and X. Shi are with Tsinghua-Berkeley Shenzhen Institute, Tsinghua University, Shenzhen, Guangdong, China, Corresponding author: Guibin Chen, e-mail: chengbits@yeah.net

charging expense while satisfy users' comfort, [17] formulates the EV charging problem as a CMDP and solves it by directly using an on-policy safe DRL algorithm, the constrained policy optimization (CPO).

Most of existing methods optimize the multiobjective tradeoff problem by adding penalties to the reward function, in which the reward function contains practical charging expense objective and penalties signals for dissatisfaction of users' charging requirements. Through revising the reward function, the original constrained optimization problem is transformed as an multiobjective optimization problem. Even the aforementioned methods have obtained promising results, their performances rely heavily on the reward-penalty ratio. Although one can find the ratio that leads to the optimal policy, it requires trial and error and varies on different tasks. A standard and well-suited formulation to solve the constrained EV charging scheduling problem is via CMDP framework [17]. While existing work using on-policy safe DRL requires a tremendous amount of data, which is not realistic and low-efficiency. Accordingly, the limitations of existing DRL-based methods for EV charging scheduling can be summarized as follows:
1) The suboptimal charging scheduling performance caused by discretization of continuous charging action;
2) The excessively reliance on penalty coefficient in revised reward function, which requires to be identified in a trail-and-error manner and is only appropriate for certain condition;
3) The on-policy safe DRL is sample inefficient.

In this paper, instead of manually designing a penalty in reward function, we formulate the optimal EV charging scheduling problem as a CMDP and propose a novel safe off-policy DRL algorithm, the augmented Lagrangian soft actor critic (AL-SAC), to directly regularize the optimization problem by adding physical constraints. By using augmented Lagrangian method, we cast the CMDP problem with continuous action space into an unconstrained saddle-point optimization problem, which enables to automatically identifies the optimal solution for primal and dual variables. Compared to the popularly adopted primal-dual method in safe DRL, the augmented Langrangian method does not require strong convexity condition for examined problem and enhance the tractability with convergence guarantee. With the maximum entropy regularization, the proposed algorithm is less sensitive to hyperparameters and enable to balance the exploration and exploitation tradeoff. The algorithm adopts a double-critics networks to simultaneously estimate the future expected reward obtained by certain actions, which enables to avoid the overestimation bias and local optimal problems in DRL training. The contribution and technical advancements are summarized as follows:
1) Instead of revising the reward function of MDP via adding penalties, this paper formulates the optimal EV charging/discharging schedule problem as a CMDP, which explicitly models the EV battery SOC constraints. By synergistically combining the augmented Lagrangian method and SAC, our proposed AL-SAC algorithm achieves high solution optimality and constraints compliance;
2) The off-policy feature of the proposed AL-SAC enables it to reuse historical data for training, which makes it more sample efficient than the popularly adopted on-policy safe DRL algorithms such as the CPO;
3) Compared to the conventional optimization-based approach such as model predictive control (MPC) method, the proposed AL-SAC algorithm is model-free and is capable to learn optimal charging strategy without exact information like users' behavior and future electricity price.

The rest of the paper is organized as follows. Section II formulates the examined problem as a CMDP after briefly introducing the definition of the concept. The framework of the proposed AL-SAC algorithm is described in Section III. The experiment results and discussions are presented in Section IV, and conclusions are provided in Section V.

## II. PROBLEM FORMULATION

In this section, we first introduce the preliminaries for CMDP and then formulate the optimal EV charging and discharging scheduling problem as a CMDP.

### A. Preliminaries of Constrained Markov Decision Process

The Markov decision process (MDP) is widely adopted to formulate the sequential decision process with its predefined transition probability to model how current state-action pair influences state in next step. By allowing for inclusion of constraints that model the concept of safety, the MDP is further generalized as constrained Markov decision process (CMDP). Generally, the CMDP can be represented by a tuple, which is consisted of a state space $S$, an action space $A$, a reward function $R$, a cost function $R^c$, a transition probability $Pr$, and a discount factor $\gamma \in [0,1]$.

Under a CMDP setting, an agent learns the optimal policy by interacting with the environment at each discrete time step, $t = 1,2,...,T$. In each of time step $t$, the agent firstly observes the current state of the environment $s_t \in S$, then select an action $a_t \in A$ guided by its policy. The action adopted at time step $t$ lead to the next state according to the unknown transition probability function $Pr(s_{t+1}|s_t, a_t)$. After the transition, the agent then receives the reward $R(s_t, a_t, s_{t+1}) \subset \mathbb{R}$ and the cost $R^c(s_t, a_t, s_{t+1}) \subset \mathbb{R}$ from the environment. In a CMDP, the reward and cost are associated with each action and state pair experienced by the agent, and the safety is maintained only if the expected discounted cost is below a certain threshold.

Instead of focusing on the reward and cost associated to individual action-state pairs, the agent aims to find the control policy $\pi$ that maximizes the long-term return and safety, which are represented by the expected discounted return $J(\pi)$ and the expected discounted cost $J^c(\pi)$:

$$max_\pi J(\pi) \qquad (1)$$

s.t.

$$J^c(\pi) \leq \bar{J} \qquad (2)$$

where $\pi$ represents a mapping from a state space $S$ to an action space $A$ for a stochastic or deterministic policy. The expected discounted return is defined as $J(\pi) = E_{\tau \sim \pi}[\sum_{t=0}^{T} \gamma^t R_t]$, where $\tau$ is a trajectory or sequence of states and actions, $\{s_0, a_0, s_1, a_1, \ldots, s_{T-1}, a_{T-1}, s_t\}$. $R_t$ is the short name for $R(s_t, a_t, s_{t+1})$. The expected discounted cost under policy $\pi$ in trajectory $\tau$ is defined in the similar manner: $J^c(\pi) = E_{\tau \sim \pi}[\sum_{t=0}^{T} \gamma^t R_t^c]$, where $R_t^c$ is $R^c(s_t, a_t, s_{t+1})$ for short.

The state-action value function $Q^\pi(s, a)$ is defined as follows:
$$Q^\pi(s, a) = E_{\tau \sim \pi}[\sum_{t=0}^{T} \gamma^t R_t \mid s_0 = s, a_0 = a] \quad (3)$$
where $Q^\pi(s, a)$ denotes the expected discounted reward starting from state $s$, taking action $a$, and thereafter following policy $\pi$. The value function satisfies the Bellman equation:
$$Q^\pi(s_t, a_t) = E_{a_{t+1} \sim \pi, s_{t+1} \sim Pr}[R_t + \gamma Q^\pi(s_{t+1}, a_{t+1})] \quad (4)$$

### B. Formulating EV Charging Scheduling Problem as CMDP

In the EV charging/discharging scheduling problem, the charging device is treated as the agent which interacts with the EV and the power grid. The state of the environment includes the SOC of the EV, and the past electricity prices. In this paper, we select the past 24 hours' electricity prices as the input state. Therefore, state is further defined as $s = (SOC_t, P_{t-23}, \ldots P_t)$. Since the capacity of EV is too small to disturb the electricity prices, the EV user is viewed as a price-taker in this paper, whose charging/discharging action exert no influence on electricity price [18],[19]. The action taken by the charging device at each time step is the discharging or charging rate, which is a continuous variable and defined as $a \in [-e_{dis}^{max}, e_{ch}^{max}]$. Noted that here the $e_{dis}^{max}, e_{ch}^{max}$ represent the maximum discharging and charging power, respectively. The negative of the discharging power denotes that the EV is discharged.

As introduced previously, the transition probability $Pr(s_{t+1}|s_t, a_t)$ is influenced by current SOC, electricity prices and charging rate. To simulate the real-world scenario, we consider that the transition probability is unknown. For the succinctness of the notation, we neglect the power loss in charging and discharging process. The dynamics of the EV's battery is further modeled as $SOC_{t+1} = SOC_t + a_t$.

The goal of the charging agent is to minimize the charging cost corresponds to the time-varying electricity prices. Thus, the reward function $R_t$ could be defined as the negative of charging cost:
$$R_t = R(s_t, a_t, s_{t+1}) = -a_t * P_t \quad (5)$$

To guarantee that the EV is fully charged before departure and avoid over-discharging, the SOC should be maintained in an appropriate range. The cost function is therefore chosen as the deviation between current SOC and the target SOC or the minimum threshold of SOC:
$$R_t^c = \begin{cases} |SOC_t - SOC_{target}|, & if\ t = T \\ SOC_{min} - SOC_t, & if\ t < T\ and\ SOC_{min} > SOC_t \end{cases} \quad (6)$$
where the first line denotes the magnitude that the SOC deviates from its target SOC at the departure time $T$. The second line represents the amount of energy that is below the minimum threshold value $SOC_{min}$.

By evaluating the feedback in the form of rewards and costs defined above via past and/or future interactions with the physical environment, the charging agent tries to learn a control policy that minimizes the charging cost while satisfying the SOC deviation constraints.

### III. CONSTRAINED RL ALGORITHM VIA AUGMENTED LAGRANGIAN METHOD

In this section, we develop an innovative constrained deep reinforcement learning algorithm based on augmented Lagrangian method, thus we named it as augmented Lagrangian soft actor-critic (AL-SAC). To solve the examined problem, the adopted reinforcement learning algorithm should be sample efficient and constraints satisfied.

*Sample efficiency:* The training of DRL algorithms generally requires the agent to interact with environment to collect sufficient samples. However, collecting a tremendous amount of operation experiences for EV charging scheduling problem repeatedly in real world is not realistic. Since the off-policy DRL algorithm's learned control policy (target policy) and the policy that generates instantaneous control action (behavior policy) are different, it allows to reuse historical operational experiences. Compared to on-policy ones, off-policy DRL algorithms are much more sample efficient and appropriate for the examined problem.

*Constraints compliance:* Traditional RL methods have not considered safe exploration. While given a RL agent complete freedom is unacceptable since certain exploratory behaviors may cause physical damage. For example, in EV charging control problem, over discharging will lead to significant SOC violations in the EV causing battery damage or dissatisfaction of the driver's charging demand. Thus, it is crucial to develop a RL algorithm, which can always achieve near constraint satisfaction.

In the following subsections, we first introduce the state-of-the-art maximum-entropy based off-policy RL algorithm, soft actor-critic (SAC) [20]. Then we present the proposed constrained RL algorithm and discuss the algorithm design.

### A. Soft Actor-Critic

Actor-critic algorithms such as PPO [21], A3C [22] and DDPG [23] have been widely adopted in the DRL application. However, the first two are featured with sample inefficient, as they require samples generated by the latest policy at each gradient step. While the DDPG often suffers from hyperparameters sensitivity. To tackle the abovementioned challenges, [20] introduces the maximum-entropy concept in [24] into the actor-critic framework and develops the soft actor critic (SAC) algorithm, which outperforms aforementioned algorithms.

By combining the entropy into the value function, SAC achieves a better tradeoff on exploration and exploitation and therefore avoids the suboptimal. The entropy for a probabilistic policy at state $s_t$ is defined as $H(\pi(\cdot|s_t)) = -\sum_a \pi(a|s_t) \log \pi(a|s_t)$. Then the state-action value function of SAC is defined as:

$$Q^\pi(s_t, a_t) = E_{a_{t+1}\sim\pi, s_{t+1}\sim Pr}[R_t + \gamma(Q^\pi(s_{t+1}, a_{t+1}) + \alpha H(\pi(\cdot|s_{t+1})))] \quad (7)$$

The policy function is defined as a probability distribution $\pi(\cdot|s_t)$ in a stochastic manner as:

$$\pi(\cdot|s_{t+1}) = \frac{e^{\frac{Q^\pi(s,\cdot)}{\alpha}}}{\sum_a e^{\frac{Q^\pi(s,a)}{\alpha}}} \quad (8)$$

### B. Augmented Lagrangian SAC

Even the SAC has achieved excellent performance on a range of challenging control tasks, it can only solve the MDPs. A popularly adopted approach allowing to tackle CMDPs with DRL algorithms is revising the reward function via adding penalties associated with infeasible control over constraint. However, simply adding the product of fixed penalty coefficient and constraint violation into the reward function will lead to an infeasible or too consecutive control policy. In addition, identifying the penalty coefficient requires trial and error, which is low-efficient and time consuming. In this subsection, we propose AL-SAC by extending SAC algorithm to satisfy the operational constraints in CMDPs.

The optimal EV charging/discharging scheduling problem can be formulated as follows:

$$\max_\pi \mathcal{J}(\pi) = \mathbb{E}_{\tau \sim \rho_\phi}\left[\sum_{t=0}^T \gamma^t R_t\right] \quad (9)$$

s.t.
$$\underline{a} \leq a \leq \bar{a} \quad (10)$$
$$\mathbb{E}_{(s_t, a_t)\sim\rho_\pi}[-\log(\pi_i(a_t|s_t))] \geq \mathcal{H} \quad (11)$$
$$\mathcal{J}^c(\pi) = \mathbb{E}_{\tau\sim\rho_\phi}[\sum_{t=0}^T \gamma^t R_t^c] \leq \overline{\mathcal{J}_c} \quad (12)$$

where the objective function is to maximize the negative of charging cost. The first line of constraints denotes the action bound, in which the $\underline{a}$ and $\bar{a}$ denote the lower and upper bound, respectively. The second line of constraints is the lower bound of entropy and the third line is the upper bound for the SOC deviation.

Noted that the action bound has already been included in the action space's definition. We adopt the augmented Lagrangian method to transfer the original constrained optimization problem into unconstrained formula:

$$\max_\pi \min_{\alpha,\lambda} \mathcal{J}(\pi) + \alpha(-\mathcal{H} - \log(\pi_i(a_t|s_t))) + \lambda(\overline{\mathcal{J}_c} - \mathcal{J}^c(\pi)) + \frac{\delta_\lambda}{2}(\overline{\mathcal{J}_c} - \mathcal{J}^c(\pi))^2 \quad (13)$$

where $\alpha, \lambda$ are multipliers for entropy constraint and cost constraint, respectively. The $\delta_\lambda$ denotes the updating step size of $\lambda$, acts as the penalty coefficient in augmented Lagrangian equation here. The entropy threshold and the upper bound for the SOC deviations are determined for specific optimization problem. In traditional RL-based methodologies, the SOC deviation is penalized directly in the reward function. In other word, they treat the variables $\alpha$ and $\lambda$ as penalty coefficients. However, the values for the penalty coefficients are hard to ensure, they can be either too conservative or infeasible. To overcome such a dilemma, we adopt the augmented Lagrangian method to solve the optimization problem, which guarantees that both primal and dual variables reach their optimal values and therefore guarantees the constraint compliance.

### C. Algorithm Design

Since the proposed AL-SAC is an off-policy DRL algorithm, parameters of actor and critic networks can be updated in an iterative manner using the historical transitions. The overall framework of the AL-SAC is summarized in Algorithm 1. In each iteration, critic networks for evaluating expected discounted reward and cost are firstly trained by using stochastic gradient descent. Then the Lagrange multipliers of corresponding constraints are updated using gradient ascent and the actor network is updated in a policy gradient manner with the unconstrained Lagrangian function.

---

**Algorithm 1:** Augmented-Lagrangian SAC Algorithm
1: **Initialization:** Total episodes: $N$, Time steps per episode: $T = 24$, Replay Buffer: $\mathcal{D}$, Parameters of actor and critic networks and corresponding target copies: $\hat{\phi}, \hat{\varphi}$.
2: **for** $n = 0$ **to** $N$:
3:     **for** $t = 0$ **to** $T$:
4:         Output action $a_t$ after receiving state $s_t$;
5:         Receiving $R_t, R_t^c, s_{t+1}$ from the environment;
6:         Set $s_t = s_{t+1}$;
7:         Store transition $\{s_t, a_t, R_t, R_t^c, s_{t+1}\}$ in $\mathcal{D}$;
#Update Parameter
8:         Sample mini-batch transitions from $\mathcal{D}$;
9:         Update critic networks $\phi, \varphi$ by Eq.(16) and Eq.(19);
10:       Update Lagrange multipliers $\alpha, \lambda$ by Eq.(23) and Eq.(24);
11:       Formulate Lagrangian function as Eq.(21) and update actor network by Eq.(22);
12:       Update target networks in soft manner.
13:     **end**
14: **end**

---

*1) Value function design and training*

In order to quantify the policies, the state-action value functions $Q^\pi(s, a)$ is defined as Eq.(9). In the Lagrangian function, the $Q^\pi(s, a)$ represents the expected discounted reward and the corresponding entropy multiplier product after taking action $a$ under state $s$ with policy $\pi$. Noted that we store the tuple $\{s_t, a_t, R_t, R_t^c, s_{t+1}\}$ in the experience replay buffer $\mathcal{D}$ in each timestep and use these data for training.

We use two sets of neural networks to approximate the action-value function $Q^\phi(s_t, a_t)$ in timestep $t$. Using Bellman equation, we could approximate the current state-action value with the expectation of all possible next state and corresponding actions with $\pi$. That is:

$$Q^\phi(s_t, a_t) \approx \mathbb{E}_{s,a,r,s_{t+1}}[y_t] \quad (14)$$
$$y_t = R_t + \gamma * Q^{\hat{\phi}}(s_{t+1}, a_{t+1}) \quad (15)$$

where the $\hat{\phi}$ is the target network which is updated using soft update: $\hat{\phi} = \eta\phi + (1-\eta)\hat{\phi}$. The $\eta$ denotes the updating rate.

Hence the training for $\phi$ is to minimize the mean square error (MSE):

$$\mathcal{L}(\phi) = \mathbb{E}_{s,a,r,s_{t+1}}\left[\left(Q^\phi(s_t, a_t) - y_t\right)^2\right] \quad (16)$$

The action-state value function for the cost is designed in similar manner. We use two sets of networks to approximate the cost value function $Q_c^\varphi(s_t, a_t)$. Similarly, the current cost value could be approximated as:

$$Q_c^\varphi(s_t, a_t) \approx \mathbb{E}_{s,a,r,s_{t+1}}[y_t^c] \tag{17}$$

$$y_t^c = R_t^c + \gamma * Q_c^{\hat{\varphi}}(s_{t+1}, a_{t+1}) \tag{18}$$

where $\hat{\varphi}$ is the target network for cost value function and we update parameters of $\varphi$ by minimizing loss:

$$\mathcal{L}_c(\varphi) = \mathbb{E}_{s,a,r,s_{t+1}}[(Q_c^\varphi(s_t, a_t) - y_t^c)^2] \tag{19}$$

The cost value function networks also adopt the same soft update manner as the reward function networks.

*2) Policy function design and training*

As discussed previously, SAC adopts a stochastic policy, in which the policy $\pi$ is defined as a probability distribution $\pi(\cdot|s_t)$. Since directly optimization of a distribution is hard to implement, the policies $\pi$ is reparametrized as

$$a_\theta(s_t, \xi_t) = \tanh(\mu_\theta(s_t) + \sigma_\theta(s_t) \odot \xi_t), \xi_t \sim \mathcal{N}(0,1) \tag{20}$$

where $\mu_\theta$ and $\sigma_\theta$ is the mean and standard deviation approximated by neural networks.

The goal of the policy network is to maximize the value of Lagrangian function. To drive the primal variable $\theta$ and dual variables $\alpha$ and $\lambda$ to reach the saddle-point $(\theta^*, \alpha^*, \lambda^*)$, we adopt the primal-dual method to alternatively update these variables.

With the defined value function, the original Lagrangian function is reformulated as follow

$$L(\theta, \alpha, \lambda) = \mathbb{E}_{(s,a,r,s_{t+1})\sim D}[Q^\phi(s_t, a_t)] + \alpha[-\mathcal{H} - \mathbb{E}_{(s_t,a_t)\sim D}[\log(\pi(a_t|s_t))] + \lambda[\overline{\mathcal{J}_c} - \mathbb{E}_{(s,a,r,s_{t+1})\sim D}[Q_c^\varphi(s_t, a_t)]] + \frac{\delta_\lambda}{2}[\overline{\mathcal{J}_c} - \mathbb{E}_{(s,a,r,s_{t+1})\sim D}[Q_c^\varphi(s_t, a_t)]]^2 \tag{21}$$

We update the primal variable $\theta$ at $i$-th iteration by using gradient ascent:

$$\theta_{i+1} = \theta_i + \delta_\theta \nabla_\theta L(\theta, \alpha, \lambda) \tag{22}$$

For the dual variables $\alpha$ and $\lambda$, we update them with the gradient ascent:

$$\alpha_{i+1} = [\alpha_i + \delta_\alpha \nabla_\alpha L(\theta, \alpha, \lambda)]_+ \tag{23}$$

$$\lambda_{i+1} = [\lambda_i + \delta_\lambda \nabla_\lambda L(\theta, \alpha, \lambda)]_+ \tag{24}$$

where the $[]_+$ is the projection to non-negative real numbers.

Noted that the implementation details such as the delayed update of value function, the double-critics network are omitted here.

## IV. NUMERICAL STUDY

In this section, we evaluate the performance of proposed AL-SAC algorithm by comparing it with several benchmark methodologies with real-world electricity price.

*A. Simulation Settings*

We adopt the real-world electricity prices data from the German day-ahead market on the European Power Exchange SE [25]. The electricity price data is hourly time-varying retail price, and the time period is from October 1st, 2018, until May 1st, 2020. We select the last 175 days' data as the test data while the rest 404 days' data as the training set. To simulate real-world EV charging scenario, we set each training/testing episode as one entire day with 24 hours. The arrival time, departure time and remaining energy at arrival of each EV are assumed to follow certain distribution pattern [26-27]. For example, EV users usually drive to work in the morning and go back home in the evening. In this paper, we set the EV's arrival time and departure time as truncated normal distribution as suggested in [27]. The distribution details of the EV's information are presented in Table I. The arrival time is assumed to follow the truncated normal distribution with mean and standard deviation value as 18 and 1, which is bounded in range [15,21]. For the departure time, the mean and standard deviation are 8 and 1, respectively; and it is bounded by [6,11]. The remaining energy of the EV in arrival time follows a truncated normal distribution with mean and standard deviation value as 50% and 10% of the EV battery capacity, which is bounded by [30%, 80%]. Similar to the EV setting in [17], we consider a Nissan Leaf EV as the example, which has a maximum battery capacity as 24 kWh. In our study, the minimum threshold value of the EV's state of charge (SOC) is set as 4.8 kWh, namely 20% of the maximum capacity. While the target SOC is set as 24 kWh to guarantee that the EV is fully charged in departure. For the SOC constraints upper bound, we select $\overline{\mathcal{J}_c}$ as 0.1% of the maximum EV battery capacity. For the charging device, its maximum charging and discharging rate is set as 6 kWh. Therefore, the action $a$ can be chosen in the range [-6,6]. Noted that the negative value of the charging rate denotes the discharging action. Without loss of generality, we assume that the energy conversion efficiency during charging and discharging process are 100%.

TABLE I
DISTRIBUTION PATTERN OF THE EV

|  | Distribution | Boundaries |
|---|---|---|
| Arrival time | $\mathcal{N}(18, 1^2)$ | [15,21] |
| Departure time | $\mathcal{N}(8, 1^2)$ | [6,11] |
| Battery SOC in arrival | $\mathcal{N}(0.5, 0.1^2)$ | [0.3, 0.8] |

*B. Benchmark Methods*

We select several state-of-the-art deep reinforcement learning algorithms and the model predictive control (MPC) method as benchmark methodologies for comparison. These deep reinforcement learning algorithms include the deep deterministic policy gradient (DDPG), soft actor-critic (SAC) and constrained policy optimization (CPO). All these DRL algorithms enable to deal with the high-dimensional continuous state and action spaces. However, the DDPG and SAC cannot be directly utilized in CMDP settings. To deal with the constraints in the examined problem, we manually select the penalty coefficient in the reward function for these algorithms. Noted that the value of penalty coefficient is hard to ensure directly, which is usually evaluated by trial and error. In our study, the reward for DDPG and SAC is reshaped as $R_t - \sigma R_t^c$, in which the penalty coefficient $\sigma$ is set as 1.2 € and 0.12 € per unit SOC violation. CPO is a state-of-the-art safe RL algorithm which can be directly used for solving CMDP. According to [28], CPO enables to maintain feasibility and guarantee monotonic policy improvement at each policy iteration given a feasible policy is recovered.

Since all the DDPG, SAC and AL-SAC are off-policy DRL algorithms, samples in each training step could be stored in the replay buffer and used for training. While the on-policy DRL algorithm such as CPO requires that training samples must be generated by current policy. In this paper, the sampling size for each iteration step of CPO is set to be 256 in order to provide sufficient samples for accurate evaluating state values.

To better illustrate the effectiveness of the proposed AL-SAC algorithm, the neural network architecture, and parameters of all these off-policy actor-critic DRL algorithms are set the same. The details of neural network are presented in Table II.

TABLE II
ALGORITHM HYPERPARAMETERS

| Algorithm | Parameter | Value |
|---|---|---|
| DDPG, SAC and partially shared for CPO, AL-SAC | Hidden layer size | [256,256] |
| | Batch size | 256 |
| | Discount factor $\gamma$ | 0.995 |
| | Learning rate for network | 5e-4 |
| CPO | Hidden layer size | [64,64] |
| AL-SAC | Learning rate for multipliers $\alpha, \lambda$ | 1e-5 |
| | Initial value for $\alpha, \lambda$ | 0 |

To illustrate the convergence and effectiveness of all these DRL algorithms, we also adopt one optimization-based algorithm. The model predictive control (MPC) is used to solve the optimal EV charging/discharging scheduling in each day. Noted that for the MPC method, the forecasted future electricity prices, and the distribution pattern of EV's departure time are given. Based on the predictions, MPC solves the problem over a planning horizon of 24 hours. We assume that the forecasted electricity price has a 10% error from that of the practical price, and the forecasted price data is generated by adding noise from normal distributions $\mathcal{N}(0, (0.1P_t)^2)$, $t = 1,2, ... T$, where $P_t$ is the practical price at time step $t$. The departure time is predicted by drawing a sample from the known distribution. The results of MPC method under ideal scenario are also presented for providing the "Theoretical-Optimum" results. Under the ideal scenario, the MPC knows accurate future electricity prices and exact departure time of the EV. The commercial solver MOSEK is adopted for the MPC method.

*C. Numerical Results*

To verify the convergence and efficiency of the proposed AL-SAC, we conduct the EV charging/discharging scheduling experiment using all these methods in the training datasets. Fig.1-2 present the trend of charging cost in training process of all these DRL algorithms as well as the "Theoretical-Optimum" value of MPC method. The SOC constraint violation value in training process of all these methods are presented in Fig.3-4.

As shown in these figures, as an on-policy DRL algorithm, CPO requires much higher number of samples for achieving a stable performance compared to off-policy DRL algorithms such as AL-SAC, SAC and DDPG. In terms of the SOC violations, the CPO converges after training with 60000 samples, while AL-SAC achieves almost the optimal performance with less than 1000 training samples. The results demonstrated that AL-SAC has considerable superiority in sample efficiency compared with CPO.

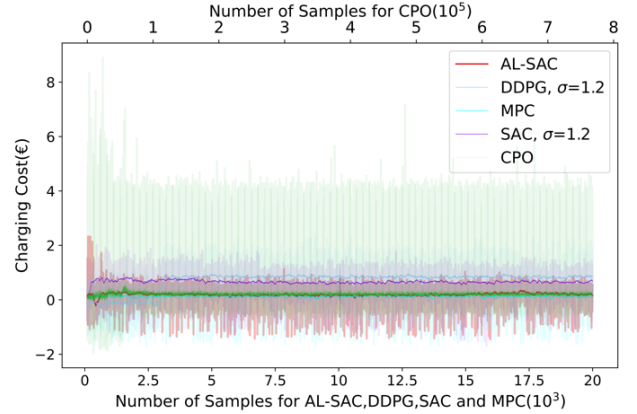

**Fig. 1.** Episode charging cost in training process.

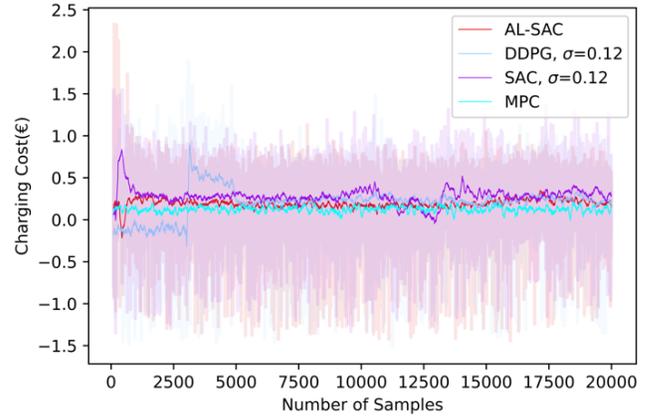

**Fig. 2.** Episode charging cost in training process.

The stable stage of all these benchmark methods in training process indicates their capability of solving the examined problem. The charging cost of the proposed AL-SAC approximates that of the MPC method, which represents the theoretical optimal value under complete information setting. As for the SOC constraints violation, the AL-SAC satisfies the SOC constraints almost all the time after converging to the stable stage. Although the CPO achieves a similar charging cost as that of the AL-SAC, it suffers relatively higher SOC constraints violations. The results indicate that CPO is not always effective in solving CMDP.

In fact, the similar results were discussed by [29] that CPO's feasibility is rarely guaranteed due to errors in gradient and Hessian matrix estimation. The performances of SAC and DDPG are significantly influenced by its penalty coefficient. With a larger penalty coefficient value ($\sigma = 1.2$), the charging cost of both DDPG and SAC are much higher, but the SOC constraints are strictly satisfied. When the penalty coefficient value decreases to $\sigma = 0.12$, the charging cost of SAC and DDPG also decrease sharply, while the SOC constraints satisfaction cannot be guaranteed. The experiment results illustrate that the proposed AL-SAC is efficient and reliable since it can adaptively optimize the objective while satisfy the constraints synchronously.

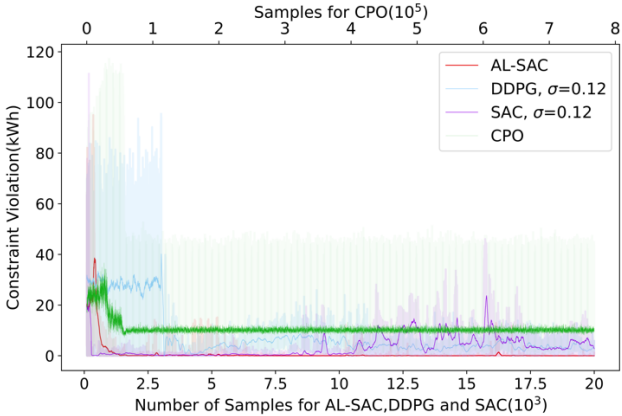

**Fig. 3.** SOC violations in training process.

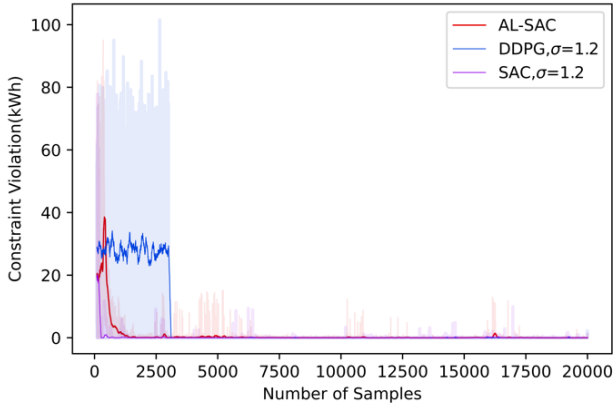

**Fig. 4.** SOC violations in training process.

Table III presents the performances of the proposed AL-SAC and other benchmark methods on the test dataset, which includes 175 days. Results of all these RL algorithms are obtained by their saved models that have best performance on training dataset.

As shown in this table, the proposed AL-SAC algorithm achieves the lowest average charging cost among all these methods except for MPC approach that under complete information setting. However, when the MPC approach can only evaluate the departure time of the EV from the distribution and the prediction error occurs in electricity price, its average charging cost increases and charging demand violations also occur frequently. Only the AL-SAC, MPC and SAC with larger penalty coefficient achieve the strict constraints compliance. However, for SAC, the larger $\sigma$ induces much higher charging cost, which reflects the superiority of our proposed AL-SAC.

TABLE III
PERFORMANCE COMPARISON ON TEST DATASET

| Algorithm | Average charging cost (€) | Average SOC violation (kWh) |
|---|---|---|
| DDPG ($\sigma$ =0.12) | 0.216 | 15.061 |
| SAC ($\sigma$ =0.12) | 0.283 | 0.389 |
| CPO | 0.163 | 10.490 |
| **AL-SAC** | **0.113** | **0.0** |
| MPC (ideal scenario) | 0.068 | 0.0 |
| MPC (10% error) | 0.120 | 0.274 |
| DDPG ($\sigma$ =1.2) | 0.805 | 0.639 |
| SAC ($\sigma$ =1.2) | 0.697 | 0.0 |

To demonstrate the superior performance of the proposed methodology, the details of charging and discharging schedules controlled by the AL-SAC are further presented in Fig.5 and Fig.6. Fig.5a presents the hourly charging and discharging rate and electricity prices in 7 consecutive days, which is randomly selected from the testing dataset. The red bar represents the charging or discharging rate controlled by the AL-SAC algorithm, which intelligently charges when the electricity price is relatively lower and discharges when the price increases within the EV's parking period. Noted that the blank area between two consecutive charging/discharging period represents the period when EV is out of home.

The corresponding EV battery capacity is shown in Fig.5b. Under the charging/discharging control of the proposed algorithm, the EV's battery is always fully charged in departure time and above the minimum SOC threshold value. The experimental results show the effectiveness of the proposed AL-SAC algorithm in optimizing real-time EV charging/discharging schedules under the battery SOC constraints.

We further conduct experiment under more complex electricity price patterns to verify the efficiency of AL-SAC. The performance of AL-SAC under 10 consecutive days is shown in Fig.6. It can be observed that AL-SAC is still stable under a totally different electricity price pattern and has no constraints violation while minimizing the charging cost.

## V. CONCLUSION

A model-free safe DRL algorithm is proposed to optimize the real-time EV charging and discharging schedules without accurate information on the EV's arrival time, departure time, remaining energy, and real-time electricity price. To satisfy the EV charging demand constraints, we formulate the optimal EV charging scheduling problem as a CMDP and consider the uncertainties in real world. We innovatively developed the constrained reinforcement learning method by introducing the augmented Lagrangian method into the soft actor-critic, which enables DRL methods to effectively handle with constrained optimization problem and ameliorate training robustness. Numerical results demonstrate that the AL-SAC outperforms the benchmark methods in solution optimality and constraints compliance.

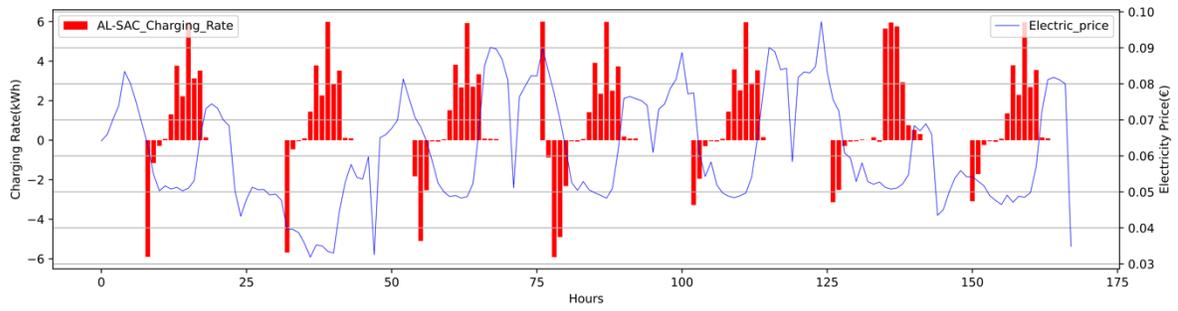

(a) Hourly electricity price (blue line) and EV charging/discharging schedules (red bar)

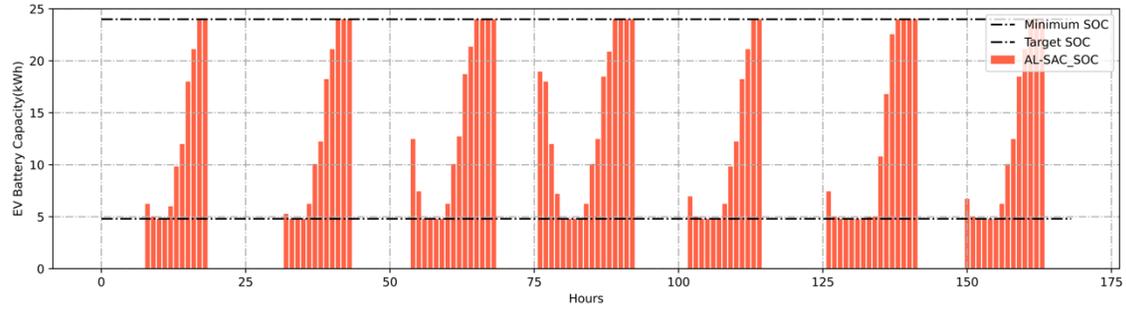

(b) Hourly SOC of EV (orange bar)

**Fig. 5.** Dynamic scheduling results achieved by AL-SAC algorithm in 7 consecutive days in test dataset.

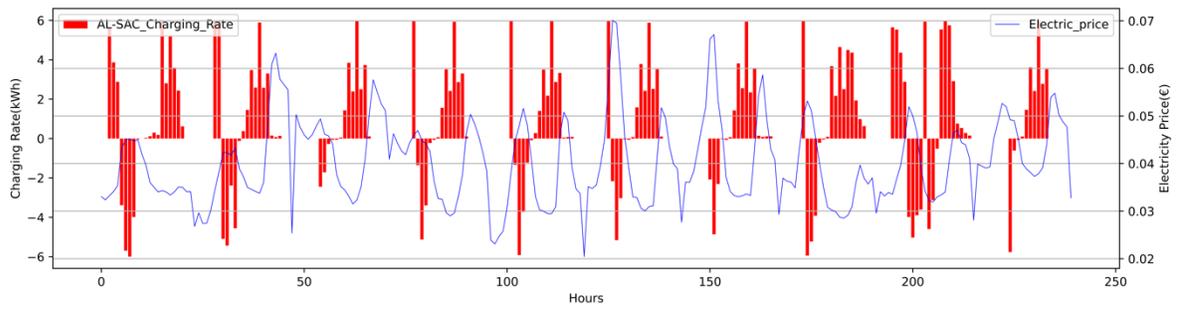

(a) Hourly electricity price (blue line) and EV charging/discharging schedules (red bar)

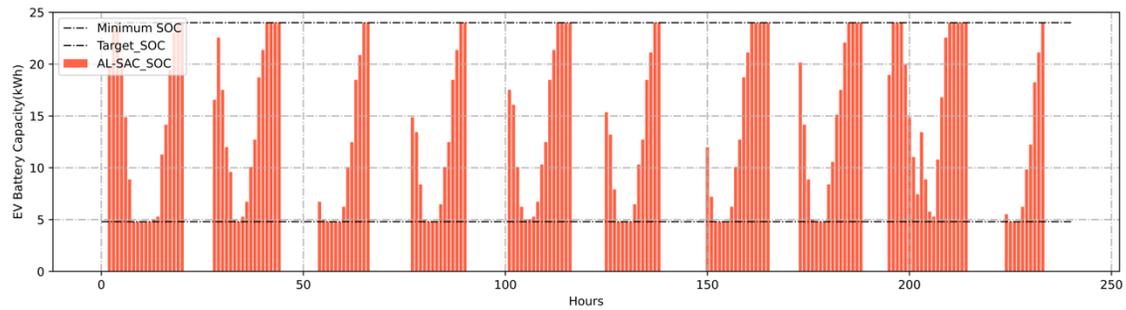

(b) Hourly SOC of EV (orange bar)

**Fig. 6.** Dynamic scheduling results achieved by AL-SAC algorithm in 10 consecutive days in test dataset.